\documentclass[runningheads]{llncs}

\usepackage{eccv}

\usepackage{eccvabbrv}

\usepackage{graphicx}
\usepackage{booktabs}
\usepackage{amsmath}
\usepackage[accsupp]{axessibility}  

\usepackage[breaklinks,colorlinks,citecolor=eccvblue]{hyperref}

\usepackage{orcidlink}

\newtoggle{final}

\usepackage{xspace}
\newcommand{\ours}{PAD-HOI\xspace}

\begin{document}

\title{Policy-as-Data: Learning Generalizable HOI Diffusion Models from Simulated Physics} 

\titlerunning{Abbreviated paper title}

\author{
Shujia Li\inst{1,2} \and
Jianshu Hu\inst{1} \and
Haiyu Zhang\inst{3} \and
Yunpeng Jiang\inst{1} \and
Haoyuan Jin\inst{1} \and
Xinyuan Chen\inst{2} \and
Yaohui Wang\inst{2} \and
Yutong Ban\inst{1}
}

\authorrunning{S.~Li et al.}

\institute{
Global College, Shanghai Jiao Tong University \and
Shanghai AI Laboratory \and
Beihang University\\
\url{https://github.com/etach-qs/PAD-HOI-Retargeter}
}

\maketitle

\begin{abstract}
Synthesizing realistic Human-Object Interactions (HOI) is critical for creating embodied avatars and functional virtual environments.
However, current data-driven approaches primarily rely on motion capture datasets, which are expensive to scale and limited in functional diversity.
Models trained with these datasets fail to
generalize to unseen objects and maintain physical consistency over long horizons.
In this paper, we propose a novel framework that leverages a physics simulator to overcome the data-scarcity bottleneck in HOI generation.
Specifically, we propose a scalable pipeline, called \ours, which leverages policies trained with reinforcement learning in a physics simulator for task-oriented data generation and trains a generative model on the augmented dataset for generalizable HOI generation.
To seamlessly utilize the synthetic data, we introduce a coarse-to-fine retargeting process that bridges the representation gap between the simplified model used in physics simulator and the standard parametric body models required for generative training.
Validated through comprehensive experiments, our method demonstrates enhanced generalization to unseen objects and the capability of long-horizon generation, while exhibiting greater dynamic diversity and physical plausibility.

\end{abstract}

\section{Introduction}
\label{sec:intro}

\begin{figure}[t]
    \centering
  
    \includegraphics[width=\textwidth]{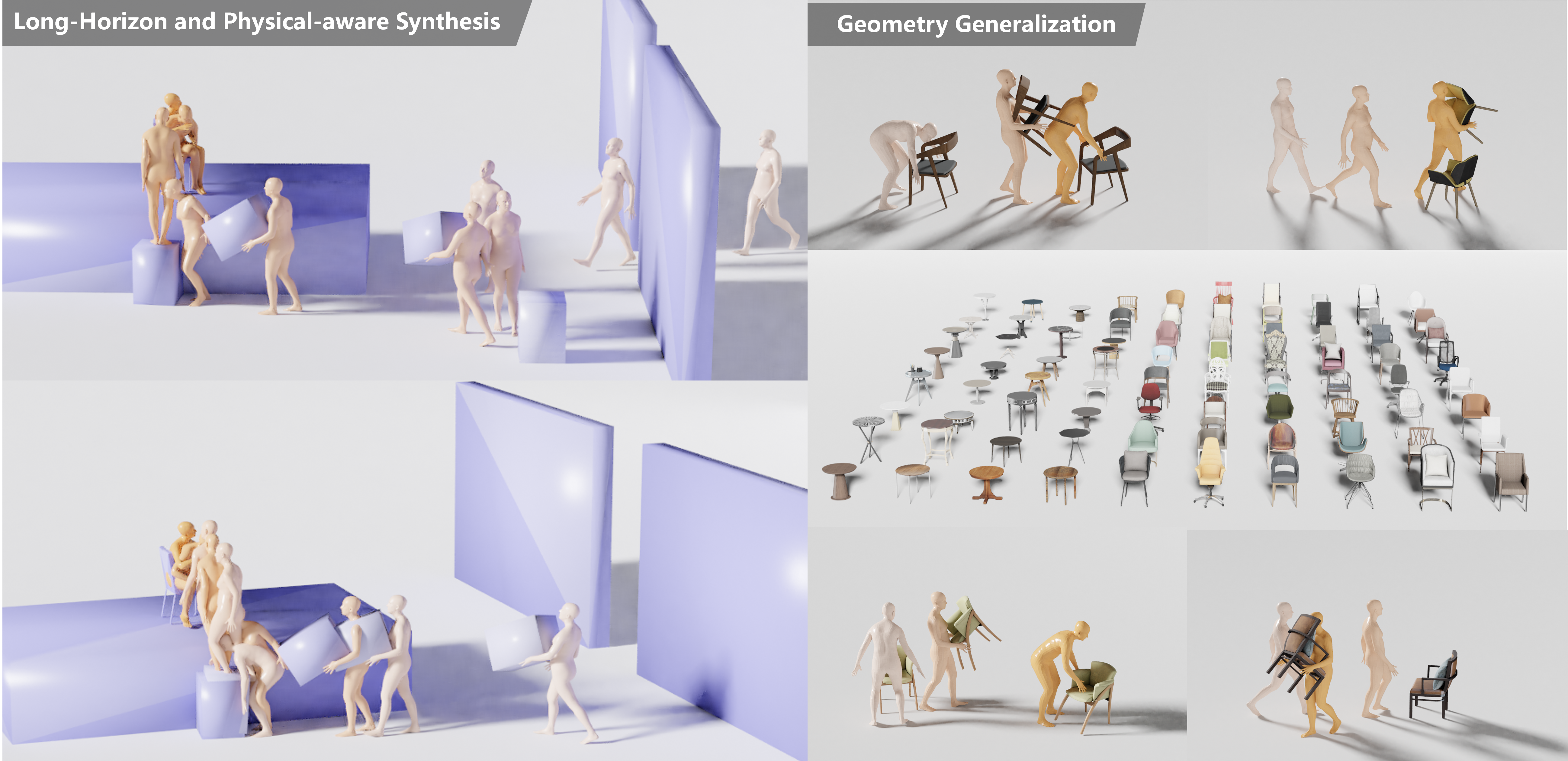}
    \caption{\textbf{Physically Grounded Long-Horizon HOI Generation.} Our framework synthesizes complex, multi-stage interactions (e.g. carrying a box and seamlessly transitioning to sit) while maintaining stable physical contact. The method dynamically adapts to diverse geometries, demonstrating robust interaction generation across over 100 distinct objects.}
    \label{fig:teaser}
\end{figure}

In daily life, humans seamlessly execute long-horizon interactions with diverse objects, such as carrying a wide box to a target location before transitioning to sit on a chair.
These multi-stage tasks require precise, geometry-aware coordination between the human body and hands to dynamically adapt postures and maintain stable contact.
Synthesizing such realistic, physically grounded Human-Object Interactions (HOI) is a foundational challenge, essential for advancing embodied AI, high-fidelity simulators, and immersive virtual environments.

Recent advancements in generative models, such as diffusion models~\cite{ho2020denoising}, have demonstrated remarkable capabilities across the image~\cite{rombach2022high,saharia2022photorealistic, esser2024scaling}, video~\cite{ho2022imagen,blattmann2023stable,bar2024lumiere}, and 3D~\cite{zhang20244diffusion,poole2022dreamfusion,yi2023gaussiandreamer} domains.
However, their potential in HOI synthesis remains constrained by severe data scarcity and physical inaccuracies of available motion capture (MoCap) datasets.
Prohibitively expensive and sophisticated acquisition pipelines restrict the scaling of MoCap datasets to diverse object geometries and long-horizon functional interactions.
Furthermore, 
capture hardware and post-processing inevitably introduce physical artifacts, such as penetration and floating.
Consequently, models~\cite{li2023object, li2024controllable} trained exclusively on these datasets overfit to a narrow range of standardized objects and exhibit physical inconsistencies, and fail to generalize to unseen object geometries or execute long-horizon tasks.

To bypass the limitations of MoCap acquisition, an alternative paradigm relies on physics simulators. Policies trained using reinforcement learning (RL) and domain randomization can generate 
task-oriented trajectories.
Specifically, pioneering works~\cite{peng2021amp, InterPhysHassan2023, gao2024coohoi} include style rewards, enabling the synthesis of natural and physically plausible interactions.
TokenHSI~\cite{pan2025tokenhsi} facilitates long-horizon functional interactions through specialized training protocols.
However, developing a universal RL controller that generalizes across varying object geometries are restricted by the complexities of RL training and reward engineering.
In contrast, generative models such as diffusion models are better suited to capturing the distribution of large-scale datasets.
A natural solution is to leverage task-specific RL experts as data generators for diffusion models.
However, this introduces a critical hurdle: \textit{representation gap}.
To maintain training efficiency and feasible policy convergence, low-DoF rigid-body representations are used in RL training.
These simplified skeletons fundamentally lack the high-DoF articulation and dense surface geometry required to represent rich, expressive human movements.
Consequently, these physics-based data cannot be directly integrated with high-fidelity MoCap datasets, which utilize the SMPL format~\cite{loper2023smpl,pavlakos2019expressive}.
To unleash the power of data generated in simulator, a robust retargeting process is required to bridge this gap.

Driven by these necessities, we propose Policy-as-Data for HOI generation (\ours), a framework that leverages a coarse-to-fine retargeting process to process physics-based simulation data into large-scale training data suitable for generative model.
Our core insight is that the physically valid, long-horizon capabilities of simulator-trained policies can be leveraged for scaling the training of diffusion models.
By augmenting limited real-world MoCap data with a comprehensive, retargeted dataset, we enable training a conditional generative model with strong geometric and physical priors.
Finally, our model can generalize to varying objects and synthesize contact-stable, and physics-valid long-horizon interactions as shown in Fig.~\ref{fig:teaser}.
In summary, our main contributions are as follows:
\begin{itemize}
    \item \textbf{The Policy as Data Framework:} We introduce a novel framework that mitigates the scarcity and limited quality of MoCap data by using task-oriented policies trained in a physics simulator as an automated data engine.
    \item \textbf{A Coarse-to-Fine Retargeting Module:} 
    We propose a coarse-to-fine retargeting module that bridges the representation gap between simulated physics and kinematic models. This two stage process seamlessly translates reduced rigid body trajectories into high fidelity parametric meshes while preserving exact surface contacts.
    \item \textbf{Generalization and Physical Fidelity:} 
    Through extensive experiments, we demonstrate that \ours achieves superior generalization to novel object geometries and strong ability of synthesizing physics-aware long-horizon interactions.
\end{itemize}

\section{Related Work}
\label{sec:related_work}
\noindent \textbf{Data-driven Interaction Synthesis}
Recent advancements in generative modeling have enabled the synthesis of realistic human-object interactions conditioned on text or object states~\cite{li2025genhoi, peng2023hoi, wu2024thor, li2023object, yang2024fhoifinegrainedsemanticaligned3d, zeng2025chainhoi, cong2025semgeomo,lu2025choice,lou2025zero,diller2024cg}. Early approaches focused on synthesizing human motion in static 3D scenes~\cite{hassan2019resolving,hassan2021stochastic,huang2023diffusion,wang2022humanise,zhang2022couch,yi2024generating}.
With the advent of diffusion models, methods like OMOMO~\cite{li2023object} and CHOIS~\cite{li2024controllable} have demonstrated the ability to generate diverse interactions with dynamic objects by learning from motion capture datasets~\cite{bhatnagar2022behave, li2023object, huang_intercap_2024, fan2023arctic, jiang2023full,zhang2022couch,liu2025core4d,geng2025auto}. However, these kinematic-based approaches suffer from two fundamental limitations. First, they lack explicit physical constraints, often resulting in artifacts such as hand-object penetration, foot sliding, or floating contacts. Second, because they are trained on fixed Mocap datasets with limited object instances, they struggle to generalize to objects with unseen geometries or scales. In contrast, our approach leverages physics-based simulator to enforce contact stability and introduces domain randomization to achieve robustness to novel object shapes.

\noindent \textbf{Physics-Based Character Control.}
Physics-based simulators offer a rigorous framework for synthesizing functionally valid motion. Reinforcement learning has been successfully applied to train characters that can perform complex tasks, ranging from locomotion~\cite{peng2021amp, peng2018deepmimic, peng2022ase, tessler2023calm, tevet2024closd, MimicKitPeng2025} to dexterous manipulation~\cite{tessler2025maskedmanipulator, luo2024omnigrasp}. 
A prominent line of research focuses on physics-aware motion imitation, developing unified controllers capable of tracking large-scale reference motions~\cite{wu2025uniphys, Luo2022FromUH,Luo2022EmbodiedSH,Luo2021DynamicsRegulatedKP, hoifhli, wang2025skillmimic, yu2025skillmimic, luo2023perpetual} or establishing versatile motion manifolds~\cite{peng2022ase, dou2022case, luo2024universal} for reusable skills. 
While these imitation-based policies excel at accurately reproducing existing kinematic data within a simulator, they are fundamentally constrained by their reliance on dense reference trajectories. Consequently, they are ill-suited to act as generators for \textit{unseen} object geometries or novel functional tasks where ground-truth motion capture does not exist.
To overcome this limitation recent works like TokenHSI~\cite{pan2025tokenhsi, xiao2024unified} and UniHSI~\cite{xiao2024unified} propose task oriented goal conditioned controllers. By utilizing adversarial motion priors to encourage naturalism these policies can autonomously discover and synthesize physically plausible manipulation strategies for diverse object shapes. However while these methods guarantee physical plausibility they typically operate on simplified reduced order models to ensure simulator stability. This structural gap limits their direct applicability to modern avatar creation pipelines which require high fidelity parametric meshes~\cite{loper2023smpl,romero2022embodied,pavlakos2019expressive} and fine grained text controllability. Our work bypasses this architectural limitation entirely by separating the physical simulation from the final generative output.

\noindent \textbf{Physics-Guided Motion Synthesis}
Bridging kinematic generation and physics simulator is an active frontier~\cite{lin2025,deng2026humanobject,PDP24}. Currently, most frameworks treat simulator merely as an auxiliary physical filter. Methods such as PhysDiff~\cite{yuan2023physdiff} and POMP~\cite{ji2025pomp} rely on physics engines for post stage motion projection or dynamically consistent denoising. Other approaches, including RLPF~\cite{yue2025rl} and PARC~\cite{xu2025parc}, utilize reinforcement learning fine-tuning and iterative physics-based correction, while InterPhys~\cite{InterPhysHassan2023} and InterAct~\cite{xu2025interact} perform post-hoc optimization to resolve contact artifacts in noisy motion capture data. Similarly, retargeting techniques like AvatarPoser~\cite{jiang2022avatarposer} map simulated states to kinematics but are typically applied to track existing trajectories rather than for autonomous dataset expansion. We propose a fundamentally different paradigm, utilizing the simulator as an upstream \textit{generative source}. Instead of correcting motions downstream, our synthetic-to-real pipeline randomizes object properties to synthesis large-scale, physically valid interactions. This directly equips the generative model with functional priors for generalizable long-horizon synthesis, eliminating the need for complex inference-time physics projection.

\section{Preliminaries}
\label{sec:preliminaries}

\noindent \textbf{Kinematic Motion Representation.} 
To provide a compact and expressive representation of human motion, we utilize the SMPL parametric body model~\cite{pavlakos2019expressive} when training the generative model. We denote a human-object motion sequence as $\mathbf{X}=[(\boldsymbol{x}^h_1,\boldsymbol{x}^o_1),...,(\boldsymbol{x}^h_T,\boldsymbol{x}^o_T)] \in \mathbb{R}^{T \times d}$, where $T$ is the sequence length and $d$ is the total feature dimension.
Here, $\boldsymbol{x}^h_t$ and $\boldsymbol{x}^o_t$ represent the human pose and object pose at timestep $t$, respectively.
The human pose $\boldsymbol{x}^h_t \in \mathbb{R}^{d_h}$ consists of the root pose (3D position $\boldsymbol{p}^h_t$ and 6D orientation $\mathbf{R}^h_t$) and the local body joint rotations $\boldsymbol{q}_t^h \in \mathbb{R}^{d_q}$, where $d_q$ represents the total joint rotation dimensions ($J$ joints $\times$ 6 dimensions per joint).
Similarly, the object pose $\boldsymbol{x}^o_t=[\boldsymbol{p}_t^o,\mathbf{R}_t^o] \in \mathbb{R}^{d_o}$ contains its global 3D centroid $\boldsymbol{p}_t^o$ and 6D orientation $\mathbf{R}_t^o$.
Specifically, $J=21$, which yields $d_q=126$, $d_h = 135$ and a total state dimension of $d = 144$.



\noindent \textbf{Physics-Based Character Representation.} 
While the SMPL model provides the expressive geometry required for high-fidelity generation, directly utilizing its highly articulated skeleton within a physics simulator is computationally intractable.
To ensure robust policy learning and scalable rollout generation, we abstract the dense SMPL geometry during simulation and instead employ a simplified physics-based humanoid character~\cite{pan2025tokenhsi}. 
We denote a trajectory $\mathbf{S} = [\mathbf{s}_1,\mathbf{a}_1,...,\mathbf{s}_T]$, containing sequences of state $\mathbf{s}$ and action $\mathbf{a}$.
Modeled as a system of $N_b=15$ rigid bodies, the character is represented by a proprioceptive state $\mathbf{s} \in \mathbb{R}^{d_s}$, including the local positions, 6D rotations, linear velocities, and angular velocities of the rigid bodies, resulting in $d_s = 222$. 
This character reduces the action space to $d_a=32$ degrees of freedom (DoF) across $N_s$ 3-DoF spherical and $N_r$ 1-DoF revolute joints, where $N_s=10$ and $N_r=2$. 
Unlike the kinematic representation which directly outputs absolute poses, our physics-based character is controlled dynamically by specifying the target rotations provided to PD controllers.

\section{Method}
\begin{figure}[t!]
    \centering
    \includegraphics[width=0.9\linewidth]{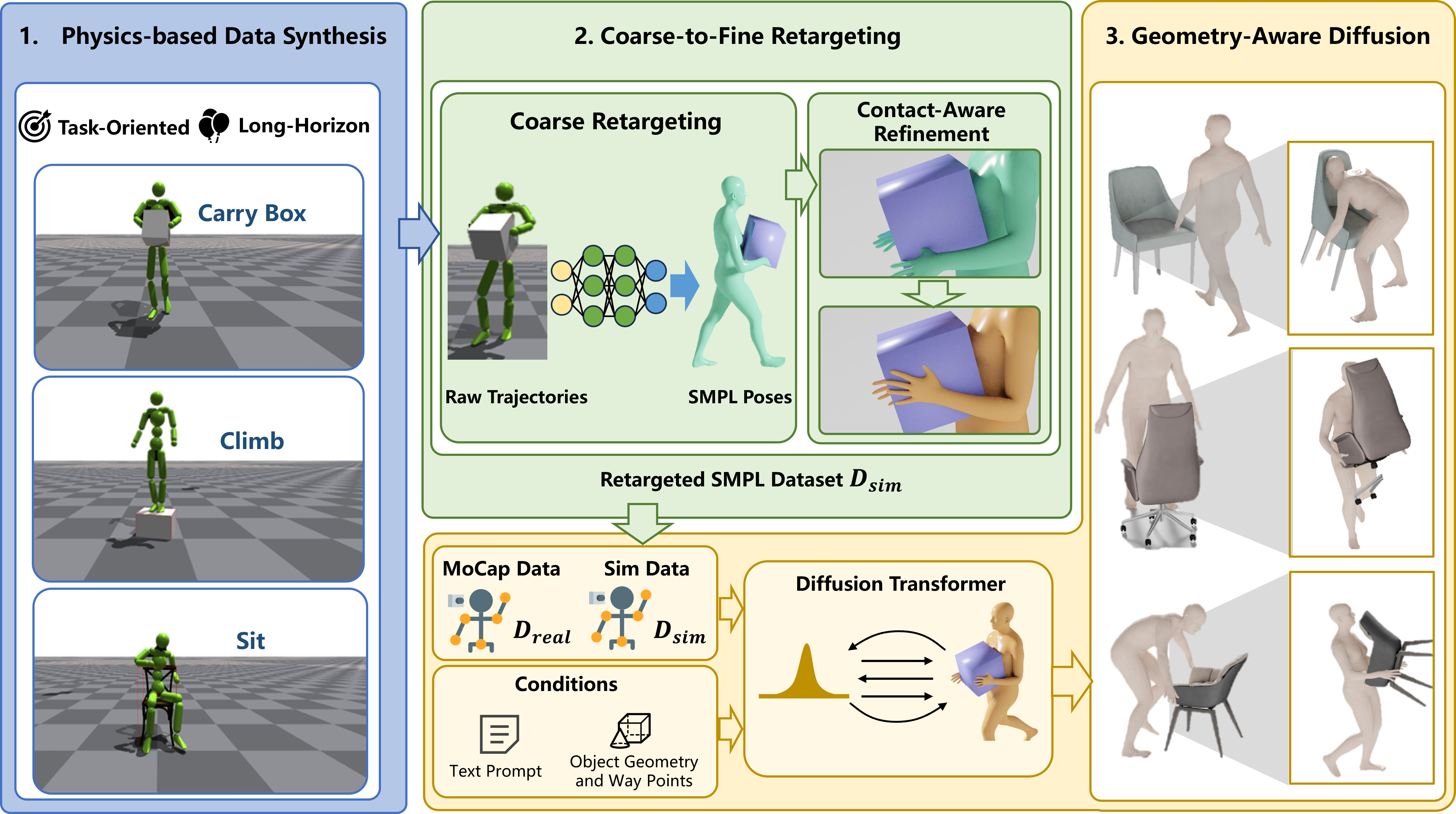}
    \caption{Overview of \ours. Our paradigm uses physics simulator to overcome MoCap data scarcity. \textbf{(a) Physics-Based Data Synthesis:} RL experts interact with procedurally randomized object geometries in simulator, generating a massive dataset of physically valid trajectories. \textbf{(b) Coarse-to-Fine Retargeting:} A retargeting module translates the simulator's rigid-body states into the high-fidelity SMPL pose space to bridge the domain gap. \textbf{(c) Geometry-Aware Diffusion:} A conditional diffusion model trained on this dataset distills the physical priors, enabling generalizable, contact-stable HOI synthesis for diversity objects.}
    \label{fig:method_pipeline}
\end{figure}

To overcome the scarcity and limited quality of real-world HOI datasets (e.g., MoCap datasets), we propose \textbf{\ours}, a novel Policy-as-Data framework, as illustrated in Figure~\ref{fig:method_pipeline}.
Our framework contains three key components: 1) a physics-based data augmentation module leveraging trained policy for trajectory generation,
2) a coarse-to-fine retargeting process to bridge the gap between expert policy rollouts and data formats required for diffusion training, and 3) a geometry-aware diffusion to learn the distribution over motion sequences from hybrid dataset.
Ultimately, our goal is to learn a generalizable diffusion model capable of synthesizing physically consistent and long-horizon Human-Object Interactions across diverse object geometries. 

\subsection{Physics-Based Data Synthesis}
\label{sec:data_synthesis}
Real-world HOI datasets, particularly those derived from MoCap, remain prohibitively expensive and difficult to scale.
Conversely, advances in physics-based simulators and RL have enabled the training of task-oriented policies capable of transferring to real-world.
We propose leveraging these expert policies not merely as controllers, but as data generator.
By querying these expert policies across procedurally randomized environments, we generate a large-scale dataset of rigid-body trajectories, denoted as $\mathcal{D}_{\mathrm{phys}}$.
Crucially, the additional computational cost introduced by training a task-specific RL policy is acceptable.
A single expert can generate diverse interactions without human capture effort.
Moreover, purely kinematic generation models rely on geometric losses or classifier guidance to approximate physical realism, often resulting in floating contacts or volume intersections.
Conversely, our RL agent operates within a strict rigid-body simulator leading to physically valid interactions.

To ensure that synthesized interactions are both functional and anthropomorphically natural, we adopt two rewards that decouples task specification from motion style~\cite{peng2021amp, InterPhysHassan2023}.
Given $s_t$ and $a_t$ which are the character state and action at time step $t$, the reward is defined as:
\begin{equation}
  r(s_t,a_t,s_{t+1},g)
  =
  w_G\, r_G(s_t,a_t,s_{t+1},g)
  +
  w_S\, r_S(s_t,s_{t+1}),
  \label{eq:amp_reward}
\end{equation} 
where $r_G$ specifies the \emph{task} objective, e.g., reaching 
a target waypoint $g$, and $r_S$ specifies a \emph{style} objective that encourages human-like transitions.
To provide a task-agnostic motion prior, $r_S$ comes from an adversarial discriminator trained to distinguish simulated transitions from real transitions from an unstructured MoCap dataset.
The policy is then trained via reinforcement learning to maximize the expected cumulative reward, encouraging the agent to complete tasks successfully while simultaneously adopting the kinematic styles of reference MoCap data.
Detailed mathematical formulations of the adversarial objective, the rewards, and the RL algorithm are provided in the Supplementary Material.

To endow our generative model with the ability to generalize to novel object geometries, we leverage domain randomization during the data synthesis phase.
Instead of training the reinforcement learning policy on a single static instance we procedurally randomize the physical dimensions and topologies of the interaction objects within the simulator~\cite{adaptnet,pan2025tokenhsi}. Specifically, our experts interact with diverse categories, including randomized boxes, tables and chairs.
Consequently, the generated dataset inherently encompasses a broad distribution of structural variations.
By distilling this geometrically diverse data, the downstream diffusion model learns robust structural priors that allow it to adapt and synthesize stable contacts for entirely unseen geometries.

\subsection{Coarse-to-Fine Retargeting}
\label{sec:retargeting}
As introduced in \Cref{sec:preliminaries}, there is a gap between the state and action space of the physics-based simulator used for RL training and the kinematic motion used for generative model learning.
To utilize the generated rigid-body trajectories $\mathcal{D}_{\mathrm{phys}}$ for diffusion training, we must bridge this gap.
We achieve this by introducing a coarse-to-fine retargeting module that processes the raw physical rollouts to form $\mathcal{D}_{\mathrm{sim}}$ and augment the original MoCap dataset.

\noindent \textbf{Coarse Retargeting.}
Mapping the low-dimensional physics skeleton back to the high-dimensional SMPL space is an under-determined reconstruction task, as the reduced skeleton inherently discards internal joint dimensions and surface geometry.
To learn this mapping with strong kinematic priors, we construct a paired training dataset using real-world OMOMO~\cite{li2023object} motions.
Considering the correspondence between reduced joint space and high-dimensional SMPL space, we extract a subset of joints from SMPL body to build the skeleton state.

Given this paired dataset, we train a neural retargeter to map low-DoF physics rollouts back to SMPL sequences.
Given a physical rollout $\mathbf{S}$ represented by joint states, we extract features $\tilde{\mathbf{S}} \in \mathbb{R}^{T \times N_b \times d_{\mathrm{phys}}}$, where $T$ is the sequence length, $N_b = 15$ is the number of rigid bodies and $d_\mathrm{phys} = 18$ is the dimension for each joint.
Here, the features of each joint contain local rotation, global rotation, linear velocity, and angular velocity.
We convert quaternions to rotation matrices, use the continuous 6D rotation representation for stability, and compute velocities via finite differences.
Given a sliding window $\mathbf{w}_t = \tilde{\mathbf{S}}[t - W + 1 : t]$ of length $W$, our trained retargeter $\mathcal{F}_\phi$ predicts the SMPL parameters at time $t$:
\begin{equation}
  \hat{\boldsymbol{x}}_t^h = \mathcal{F}_\phi(\mathbf{w}_t)
  =
  \{\hat{\boldsymbol{p}}^h_t,\hat{\mathbf{R}}^{h}_{t},\hat{\boldsymbol{q}}^{h}_{t}\},
\end{equation} 
where $\hat{\boldsymbol{p}}^h_t$ is the global root translation, $\hat{\mathbf{R}}^{h}_{t}$ is the global root orientation, and $\hat{\boldsymbol{q}}^{h}_{t}$ stacks the local 6D rotations SMPL body joints.
$\mathcal{F}_\phi$ uses a Transformer-based temporal encoder to aggregate motion history and suppress high-frequency jitter, followed by lightweight prediction heads. We train $\mathcal{F}_\phi$ with a kinematics-driven objective that combines rotation supervision with Cartesian consistency via differentiable SMPL forward kinematics:
\begin{equation}
    \mathcal{L}_{\phi} =
    \lambda_R \mathcal{L}_{\mathrm{rot}}(\hat{\mathbf{R}}^{h}, \mathbf{R}^{h})
    + \lambda_P \mathcal{L}_{\mathrm{rot}}(\hat{\boldsymbol{q}}^{h}, \boldsymbol{q}^{h})
    + \lambda_J \|\hat{\mathbf{J}} - \mathbf{J}\|_2^2
    + \lambda_V \|\hat{\dot{\mathbf{J}}} - \dot{\mathbf{J}}\|_2^2,
\end{equation}
where $\mathcal{L}_{\mathrm{rot}}$ measures the distance between rotations, and $\hat{\mathbf{J}}$ and $\hat{\dot{\mathbf{J}}}$ denote SMPL joint positions and joint velocities obtained by applying forward kinematics to $\hat{\boldsymbol{x}}^h_{1:T}$. This coarse stage produces stable SMPL motion that preserves the original simulator dynamics, yielding a consistent kinematic representation suitable for HOI diffusion training.

\noindent \textbf{Contact-Aware Refinement.}
The mapping from a reduced physics skeleton to the full SMPL surface yields a stable SMPL sequence.
However, small mesh-level artifacts, especially around the hands during contact-rich interactions, exist in the transferred motions.
We therefore apply an offline, gradient-based refinement following~\cite{xu2025interact}.

Starting from the retargeted initialization $\{\hat{\boldsymbol{x}}^{h}_t, \hat{\boldsymbol{x}}^{o}_t\}_{t=1}^{T}$, we optimize strictly the human poses to reduce penetration and jitter while improving contact consistency.
Crucially, we keep the object trajectory $\hat{\boldsymbol{x}}^o_t$ completely frozen,
and optimize the human pose $\hat{\boldsymbol{x}}^{h}_t$ by minimizing:
\begin{equation}
    \mathcal{L}(\hat{\boldsymbol{x}}^{h}_t) = E_{\mathrm{rec}} + \lambda_{\mathrm{cont}}E_{\mathrm{cont}} + \lambda_{\mathrm{pen}}E_{\mathrm{pen}} + \lambda_{\mathrm{sm}}E_{\mathrm{sm}} +
    \lambda_{\mathrm{prior}}E_{\mathrm{prior}} ,
\end{equation}
where $E_{\mathrm{rec}}$ keeps the refined human pose close to the coarse retargeted initialization.
The remaining terms ensure physical and anatomical plausibility.
We notice that hand-object intersection artifacts are severe in the retargeted motions. To resolve this, the optimization is heavily centered on the palm vertices.
We define the contact term $E_{\mathrm{cont}}$ using simulator-provided contact indicators $c_t$ and a set of interaction vertices $\mathcal{V}_{\mathrm{contact}}$ sampled exclusively from the two palms:
\begin{equation}
    E_{\mathrm{cont}} = \sum_{t=1}^{T} c_t \sum_{j \in \mathcal{V}_{\mathrm{contact}}} d\!\left(\mathbf{v}_j(t), \mathcal{O}(t)\right),
\end{equation}
where $\mathbf{v}_j(t)$ denotes the $j$-th SMPL vertex and $d(\cdot,\mathcal{O}(t))$ is a point-to-object distance computed via the Signed Distance Field (SDF) of the object.
As for the other terms, $E_{\mathrm{pen}}$ penalizes human vertices that penetrate into the object, $E_{\mathrm{sm}}$ minimizes joint accelerations to suppress temporal jitter, and $E_{\mathrm{prior}}$ regularizes the SMPL parameters to maintain natural human articulations.
The detailed explanations about other terms are introduced in Supplementary Material considering the limited space.

\subsection{Geometry-Aware Diffusion for HOI Synthesis}
\label{sec:generation}

Augmented with the retargeted dataset, which contains diverse and physically consistent interactions, we can finally train a generalizable diffusion model.
By learning the distribution of our hybrid dataset combining synthetic trajectories $\mathcal{D}_{\mathrm{sim}}$ and real motion capture $\mathcal{D}_{\mathrm{real}}$, the model absorbs the physical rules obeyed by the simulation data while retaining the natural human kinematics of the real data.
This synergy enables the model to adapt to unseen object geometries and seamlessly execute complex multi-stage tasks.

We implement our generative model using a standard Transformer-based diffusion architecture conceptually building upon CHOIS~\cite{li2024controllable}.
To fully exploit the rich geometric and temporal variations in our hybrid dataset, the model relies on a unified conditioning context.
We encode text prompts via CLIP~\cite{radford2021learning} for semantic task guidance.
To explicitly capture varying object sizes and topologies, we extract Basis Point Set~\cite{prokudin2019efficient} features from the object shape.
We also formulate a masked motion representation containing initial kinematic states and sparse future waypoints to guide coherent generation across long horizon interactions.
By utilizing this architecture, we effectively translate the physical and kinematic priors of our hybrid dataset into a highly controllable synthesis engine.

To synthesize long-horizon interactions, we partition the target sequence into overlapping temporal windows.
Within each window, we condition the diffusion model on sparse object key-points including start and end positions alongside intermediate waypoints.
We generate the sequence autoregressively in a canonical local coordinate.
To ensure temporal continuity across boundaries, we utilize the overlapping frames from the preceding window as a bridge.
We transform the states of humanoid and object in these frames into the new local coordinate system and inject them as hard constraints during the denoising process.
Finally, we stitch the generated windows together in the global coordinate space.

\section{Experiments}
\label{sec:experiments}
In this section, we comprehensively evaluate the effectiveness of our proposed framework. First, we detail the experimental setups, including the hybrid datasets, baseline methods, and evaluation metrics. Next, we demonstrate the superior geometric generalization and generation quality of \ours on unseen objects and seen objects.
We then evaluate the ability of our framework on synthesize coherent, long-horizon interactions by seamlessly combining diverse, multi-stage skills.
Finally, we validate our coarse-to-fine retargeting module, proving its critical role in bridging the simulator-kinematic gap while maintaining high physical interaction quality.

\subsection{Experimental Settings}
\label{subsec:setup}
\noindent \textbf{Datasets.}
To evaluate the effectiveness of our proposed framework, we construct a hybrid training set combining real world motion capture, denoted as $\mathcal{D}_{\text{real}}$, with our synthesized simulator data, denoted as $\mathcal{D}_{\text{sim}}$. 
For $\mathcal{D}_{\text{real}}$, we utilize the OMOMO~\cite{li2023object} dataset to provide a robust baseline of natural human kinematics.
For $\mathcal{D}_{\text{sim}}$, we leverage our reinforcement learning experts to generate over 4400 physically grounded trajectories across three distinct multi-stage tasks.
\textbf{Carry Tasks:} This category involves approaching, picking up, and placing diverse boxes, tables, and chairs.
\textbf{Sit Tasks:} This category focuses on interacting with varying chair geometries.
\textbf{Climb Tasks:} This category features locomotion onto elevated platforms.
These raw rigid body episodes are subsequently processed through our coarse-to-fine retargeting module, detailed in Section \ref{sec:retargeting}, to bridge the representation gap and yield the final SMPL dataset used for diffusion training.
Exhaustive details regarding dataset composition, object topologies, and simulation initialization parameters are provided in the Supplementary Material.


\noindent \textbf{Implementation Details.}
We implement our complete framework, including both the Coarse-to-Fine retargeter and the diffusion model, in PyTorch and train the generative models on eight NVIDIA RTX 4090 GPUs. The physics-based trajectories are generated using the Isaac Gym simulator to ensure highly parallelized data synthesis.
More implementation details are provided in the Supplementary Material.

\subsection{Geometric Generalization Performance}
A fundamental limitation of purely data-driven HOI models is their tendency to overfit to the limited object topologies present in MoCap datasets.
To evaluate whether \ours successfully imparts robust geometric priors, we assess generation quality on both out-of-distribution (unseen) and in-distribution (seen) object geometries.
Moreover, we validate the effect of our contact-aware refinement in the retargeting module by conducting an ablation study on it.

\noindent \textbf{Baselines.}
To evaluate the efficacy of our proposed pipeline, we compare \ours against leading state-of-the-art kinematic HOI synthesis methods: InterDiff~\cite{xu2023interdiff}, MDM~\cite{tevet2023human}, CHOIS~\cite{li2024controllable} and OMOMO~\cite{li2023object}.
All baseline methods are trained on the standard OMOMO training set $\mathcal{D}_{real}$ following their official implementations.
In contrast, our method utilizes the same generative backbone as CHOIS~\cite{li2023object} but is trained on our proposed hybrid dataset ($\mathcal{D}_{real} \cup \mathcal{D}_{sim}$).
This comparison isolates the impact of our physics-based data augmentation, demonstrating whether scaling with synthetic expert trajectories benefits generalization even on real-world test sets.

\noindent \textbf{Evaluation Metrics.} Following prior work~\cite{li2024controllable}, we employ three standard metric categories to comprehensively evaluate our framework.
a) \textit{Condition Matching} ($T_s, T_e, T_{xy}$) evaluates the model's adherence to spatial waypoints.
b) \textit{Interaction Quality} ($C_{\text{prec}}, C_{\text{rec}}, C_{F1}, C_\%, P_{\text{hand}}$) assesses the physical fidelity of human-object contacts and penalizes geometric penetrations.
c) \textit{Ground Truth Difference} (MPJPE, $T_{\text{root}}, T_{\text{obj}}, O_{\text{obj}}$) measures the kinematic deviation from reference motions. Detailed mathematical formulations and unit specifications for all metrics are provided in the Supplementary Material.


\begin{table*}[t!]
    \centering
    \caption{\textbf{Quantitative comparison on unseen objects.} To evaluate zero-shot geometric generalization, all models are trained on a subset of 10 object categories and evaluated on the remaining 5 strictly unseen objects. \ours significantly outperforms the state-of-the-art CHOIS baseline in both overall contact accuracy $C_{F1}$ and MPJPE. This demonstrates that the geometric priors learned from our procedurally randomized simulator data effectively prevent the model from overfitting to training topologies, enabling robust adaptation to novel object shapes.}
    \label{tab:single_window_cmp_unseen}
    \resizebox{0.9\textwidth}{!}{
    \begin{tabular}{lccccccccc}
        \toprule
        & \multicolumn{5}{c}{Interaction Quality} & \multicolumn{4}{c}{Ground Truth Difference} \\
        \cmidrule(lr){2-6}\cmidrule(lr){7-10}
        Method & $C_{rec}\uparrow$ & $C_{prec}\uparrow$ & $C_{F1}\uparrow$ & $C_{\%}$ & $P_{hand}\downarrow$ & MPJPE$\downarrow$ & $T_{root}\downarrow$ & $T_{obj}\downarrow$ & $O_{obj}\downarrow$ \\
        \midrule
        InterDiff~\cite{xu2023interdiff} & 0.18 & 0.48 & 0.25 & 0.20 & 0.62 & 30.15 & 68.40 & 96.50 & 1.95 \\
        MDM~\cite{tevet2023human} & 0.32 & 0.58 & 0.39 & 0.33 & 0.74 & 21.60 & 42.10 & 36.20 & 2.10 \\
        HOI-Diff~\cite{peng2023hoi} & 0.45 & 0.65 & 0.51 & 0.40 & 0.76 & 19.85 & 37.55 & 28.40 & 1.75 \\
        OMOMO~\cite{li2023object} & 0.03 & 0.16 & 0.05 & 0.10 & 0.26 & 33.40 & 80.15 & 0.00 & 0.00 \\
        CHOIS~\cite{li2024controllable} & 0.59 & 0.76 & 0.64 & 0.56 & \textbf{0.13} & 16.50 & 28.78 & 14.29 & 1.04\\
        \midrule
        \textbf{\ours (Ours)} & \textbf{0.64} & \textbf{0.83} & \textbf{0.68} & \textbf{0.59} & 0.19 & \textbf{16.18} & \textbf{27.76} & \textbf{14.24} & \textbf{1.03}\\
        \bottomrule
    \end{tabular}
    }
\end{table*}

\noindent \textbf{Zero-Shot Generalization.} To rigorously evaluate adaptation to novel geometries, we partition the 15 distinct objects in the OMOMO dataset following~\cite{li2024controllable}. We train the models on 10 objects and evaluate on the remaining 5.
To ensure a strict zero-shot setting, these test objects are completely unseen and excluded from both the real world training split $\mathcal{D}_{\mathrm{real}}$ and our synthetic dataset $\mathcal{D}_{\mathrm{sim}}$.

As shown in \Cref{tab:single_window_cmp_unseen}, baseline methods suffer from substantial degradation, whereas \ours exhibits markedly stronger zero-shot adaptation.
CHOIS struggles to maintain accurate interaction and achieves a contact $F_1$ score of only 0.64.
In contrast, \ours raises $C_{prec}$ from 0.76 to 0.83, improves overall contact $F_1$ to 0.68, and reduces MPJPE.
Conditioned on continuous geometric features and trained across diverse simulated object topologies, \ours learns a robust structural prior.
This allows our framework to synthesize functional and physically accurate contacts for entirely novel shapes.

\begin{table*}[t!]
    \centering
    \caption{\textbf{Quantitative comparison on seen objects (OMOMO Test Set).} All baselines are trained on real data $\mathcal{D}_{real}$ only, while our approach uses hybrid data. \ours outperforms the CHOIS in contact quality  and human pose accuracy, demonstrating that our physically valid synthetic data improves generation fidelity even on in-distribution geometries.}
    \label{tab:single_window_cmp_seen}
    \resizebox{0.9\textwidth}{!}{
    \begin{tabular}{lccccccccc}
        \toprule
        & \multicolumn{5}{c}{Interaction Quality} & \multicolumn{4}{c}{Ground Truth Difference} \\
        \cmidrule(lr){2-6}\cmidrule(lr){7-10}
        Method & $C_{rec}\uparrow$ & $C_{prec}\uparrow$ & $C_{F1}\uparrow$ & $C_{\%}$ & $P_{hand}\downarrow$ & MPJPE$\downarrow$ & $T_{root}\downarrow$ & $T_{obj}\downarrow$ & $O_{obj}\downarrow$ \\
        \midrule
        InterDiff~\cite{xu2023interdiff} & 0.28 & 0.63 & 0.33 & 0.27 & 0.55 & 25.91 & 63.44 & 88.35 & 1.65 \\
        MDM~\cite{tevet2023human} & 0.47 & 0.72 & 0.53 & 0.43 & 0.66 & 17.86 & 34.16 & 24.46 & 1.85 \\
        HOI-Diff~\cite{peng2023hoi} & 0.62 & 0.77 & 0.64 & 0.49 & 0.69 & 17.44 & 32.28 & 22.75 & 1.51 \\
        OMOMO~\cite{li2023object} & 0.66 & 0.77 & 0.67 & 0.56 & 0.55 & 15.82 & 24.75 & 0.00 & 0.00 \\
        CHOIS~\cite{li2024controllable} & 0.64 & \textbf{0.80} & 0.67 & 0.54 & 0.59 & 15.30 & 24.43 & \textbf{12.53} & 0.99 \\
        \midrule
         \textbf{\ours (Coarse Only)} & \textbf{0.67} & 0.79 & \textbf{0.69} & 0.57 & 0.61 & \textbf{14.97} & \textbf{22.39} & 13.89 & \textbf{0.90}\\
        \textbf{\ours (Coarse + Refine)} & \textbf{0.67} & 0.79 & \textbf{0.69} & \textbf{0.58} & \textbf{0.54} & \textbf{14.97} & \textbf{22.39} & 13.89 & \textbf{0.90} \\
        \bottomrule
    \end{tabular}
    }
\end{table*}



\noindent \textbf{In Distribution Performance:} We evaluate performance on seen objects from the training data $\mathcal{D}_{real}$ to ensure in domain fidelity. As shown in \Cref{tab:single_window_cmp_seen}, \ours consistently outperforms the strongest baseline CHOIS across both interaction quality and kinematic accuracy. For example, \ours increases the $C_{F1}$ score to 0.69 and reduces the MPJPE from 15.30 to 14.97. These metrics prove that augmenting real motion capture data with synthetic simulator trajectories does not degrade natural human kinematics. Instead, the simulator data acts as a powerful regularizer. This regularization prevents the generator from collapsing into nonphysical interpolations even when interacting with familiar objects.

Ultimately, the strong zero-shot generalization on unseen objects combined with this improved in distribution performance demonstrates that \ours learns robust geometric priors. Our pipeline successfully bridges the representation gap between simulated physics and natural human motion.

\noindent \textbf{Ablation on Retargeting Quality.} To verify how retargeting quality impacts downstream generation, we train our framework on two hybrid dataset variations differing only by the contact-aware refinement step.
As shown in the bottom rows of Table \ref{tab:single_window_cmp_seen}, incorporating this refinement yields motions with higher contact fidelity and significantly lower mesh penetration.
This proves that an accurate retargeting is essential for distilling simulator physics into the generative model.
Further ablations on dataset scale and model architecture are provided in the Supplementary Material.

\subsection{Physically Grounded Interaction and Multi-Skill Composition}
Considering that the baseline models are exclusively trained on $\mathcal{D}_{real}$, evaluating them quantitatively on $\mathcal{D}_{sim}$ would constitute an unfair comparison. Instead, we qualitatively demonstrate the generative capabilities of \ours on 1) object geometries from the synthetic domains 2) long-horizon motions.

\begin{figure}[t!]
    \centering
    \includegraphics[width=0.9\linewidth]{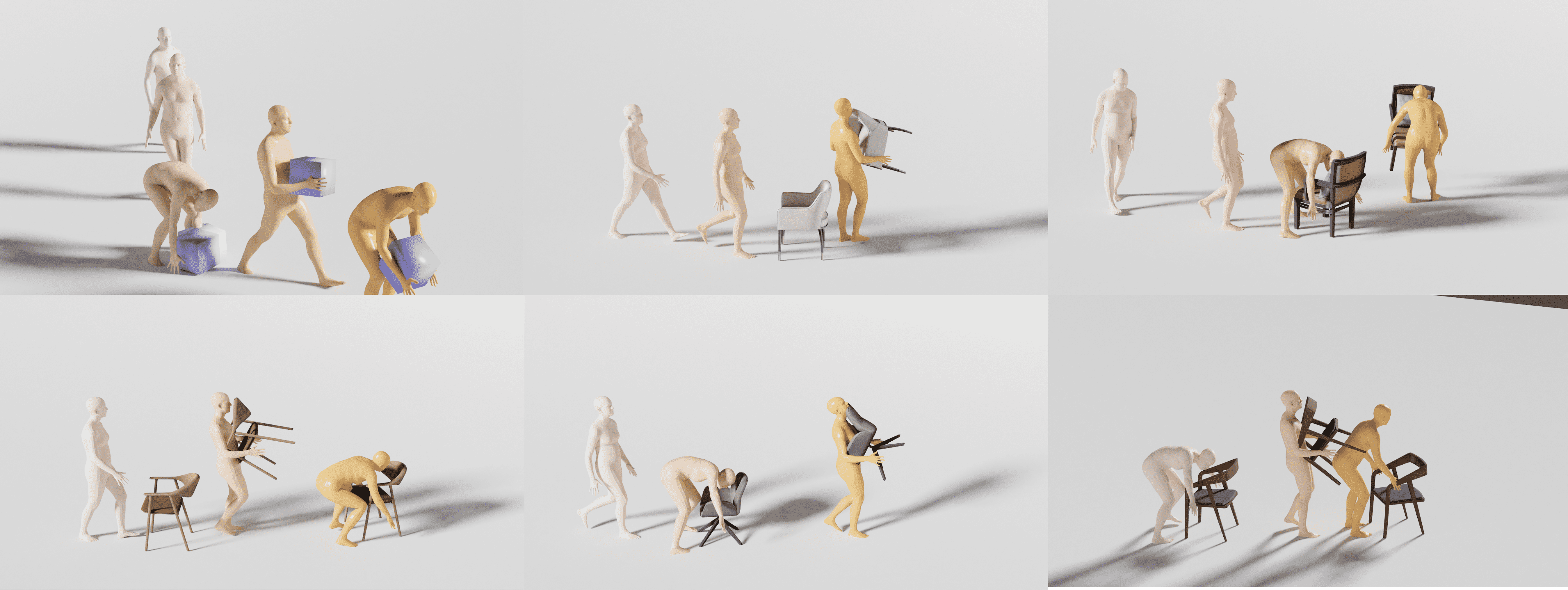} 
    \caption{\textbf{Qualitative results on 
    objects in $\mathcal{D}_{sim}$.} \ours generates highly realistic, physically plausible interactions with procedurally generated objects. The model accurately adapts the human pose to varying object shapes and scales, maintaining strict surface contacts without unnatural penetrations.}
    \label{fig:sim_qualitative}
\end{figure}

\noindent \textbf{Synthetic Domains.}
As shown in \Cref{fig:sim_qualitative}, our method synthesizes highly diverse interactions that strictly adhere to the physical constraints.
For instance, when carrying boxes, the hands of the human
press firmly against the surfaces.
Similarly, when interacting with the complex topologies like chairs, the human lifts the object by firmly gripping its sides with both arms, automatically adjusting its bimanual hold to accommodate varying widths.
Unlike standard kinematic generations that frequently exhibit floating end-effectors or intersecting meshes, \ours successfully grounds the human motion to these varying object structures, confirming that the physics-based priors from the synthetic data are effectively absorbed by our generative backbone.

\begin{figure}[t!]
    \centering
    \includegraphics[width=0.9\linewidth]{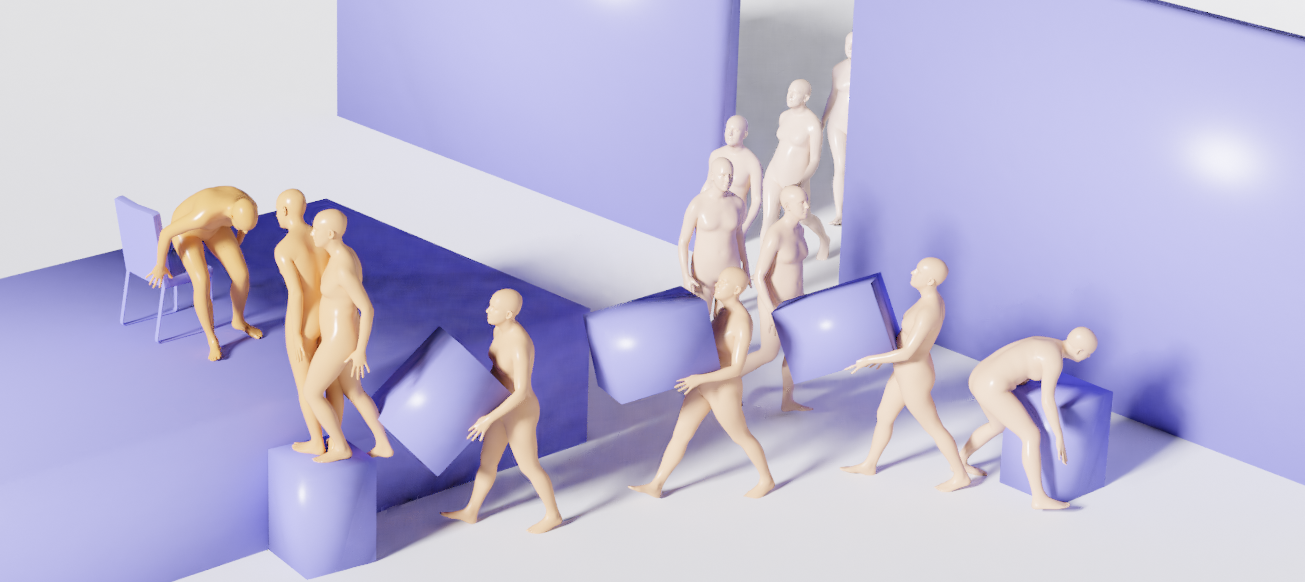} 
   \caption{\textbf{Qualitative results of multi skill long horizon generation.} Guided by object geometries spatial waypoints and the text prompt "The human navigates a narrow wall gap then approaches and picks up a box then carries it to a platform then climbs onto the box and then the platform and finally sits on a chair", \ours seamlessly chains diverse interaction skills into a coherent continuous sequence.}
    \label{fig:long_horizon_qualitative}
\end{figure}

\noindent \textbf{Long-Horizon Tasks.}
Moreover, we qualitatively assess the ability of our framework on synthesizing coherent motions of multi-stage long-horizon tasks.
Baselines trained solely on standard MoCap datasets struggle with multi-skill composition, as they lack the diverse
and physically grounded skills
required for complex scene interactions.
In contrast, by training on a rich dataset synthesized from expert policies, \ours seamlessly chains different functional skills. As shown in \Cref{fig:long_horizon_qualitative}, guided by text prompts, object geometries, and spatial waypoints, the generated avatar successfully navigates through a narrow wall gap, approaches and carries a box to a platform, climbs onto the box and then the platform, and finally sits on a chair.
This demonstrates the powerful multi-skill composition capabilities of our framework.

\subsection{Data Quality with Coarse-to-Fine Retargeting}
\label{subsec:exp_retargeting}
To generate feasible rollouts, the retargeting module (Sec.~\ref{sec:retargeting}) must maintain interaction semantics, such as accurate contacts, while avoiding mesh penetrations.
We therefore evaluate retargeting quality directly on simulator rollouts that contain diverse object geometries and contact-rich behaviors.

\noindent \textbf{Compared Methods.}
We compare three retargeting strategies.
\textbf{Mapping+IK}~\cite{aristidou2018inverse} is a heuristic baseline that maps reduced-skeleton joint targets to SMPL joints and solves per-frame inverse kinematics without explicit contact objectives.
\textbf{Coarse} is our learned retargeter, which uses a temporal window to suppress high-frequency jitter.
\textbf{Coarse+Refine} applies our offline refinement on the coarse prediction.

\begin{table*}[t!]
    \centering
    \caption{\textbf{Quantitative evaluation of coarse-to-fine retargeting strategies.} We compare our learned Coarse and Coarse+Refine methods against a standard heuristic baseline on held-out simulator rollouts. Our neural retargeter (Coarse) significantly improves kinematic tracking, while the optional offline optimization (Coarse+Refine) yields the highest interaction quality by explicitly enforcing contact constraints and resolving mesh interpenetrations.}
    \label{tab:retargeting_eval}
    
    \small 
    
    \begin{tabular}{lccccc}
        \toprule
        & \multicolumn{5}{c}{Interaction Quality} \\
        \cmidrule(lr){2-6}
        Method & $C_{prec}\uparrow$ & $C_{rec}\uparrow$ & $C_{F1}\uparrow$ & $C_{\%}$ & $P_{\mathrm{hand}}\downarrow$ \\
        \midrule
        Mapping+IK~\cite{aristidou2018inverse} & 0.50 & 0.08 & 0.13 & 0.05 & 0.00 \\
        Coarse (Ours) & \textbf{0.72} & \textbf{0.99} & \textbf{0.83} & \textbf{0.41} & 0.18 \\
        Coarse+Refine (Ours) & \textbf{0.72} & \textbf{0.99} & \textbf{0.83} & \textbf{0.41} & \textbf{0.02} \\
        \bottomrule
    \end{tabular}
\end{table*}
\noindent \textbf{Results.}
As shown in Table \ref{tab:retargeting_eval}, our Coarse method and our Coarse with Refinement method achieve identical and state of the art contact scores. The addition of the refinement stage maintains these exact high quality metrics while drastically reducing the hand penetration $P_{\mathrm{hand}}$ from 0.18 down to 0.02. This confirms that our offline optimization successfully resolves mesh interpenetrations against the object SDF without degrading the overall interaction fidelity.

\begin{figure}[t!]
    \centering
    \includegraphics[width=\linewidth]{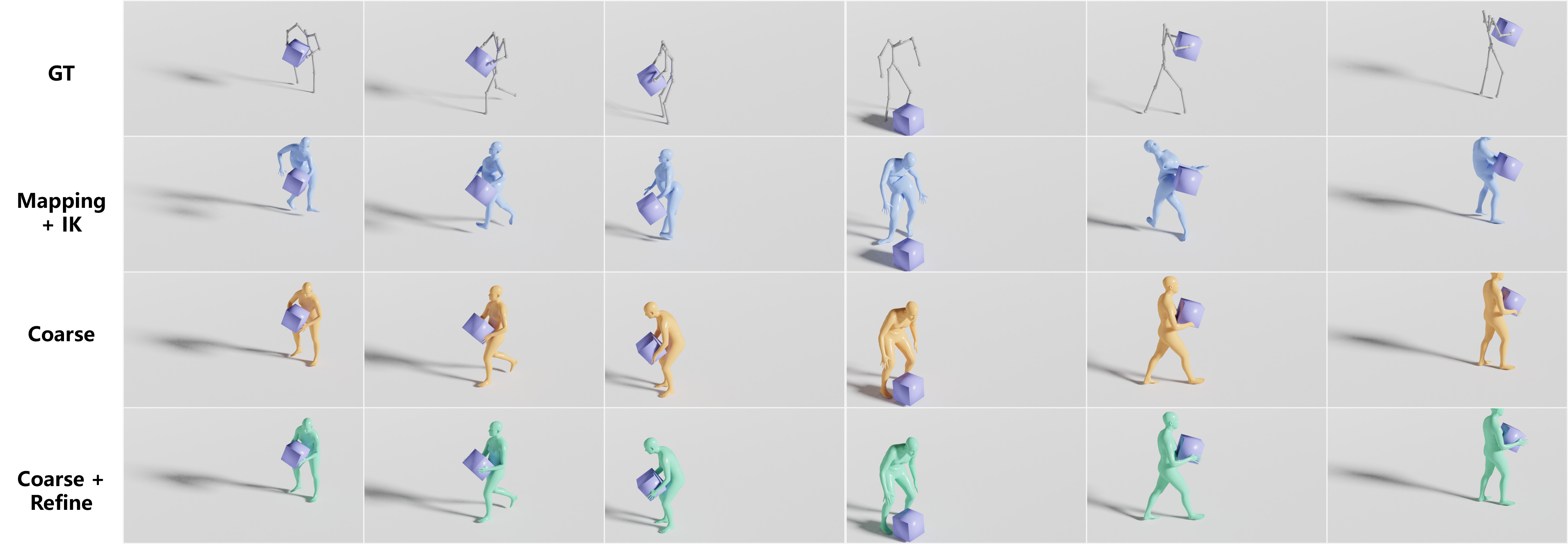} 
   \caption{\textbf{Qualitative comparison of coarse to fine retargeting strategies.} \textit{Row 1 Ground Truth:} The reference physical interaction from the simulator. \textit{Row 2 Mapping+IK:} This baseline exhibits unnatural and distorted joint configurations alongside severe mesh intersections and floating contacts. \textit{Row 3 Coarse:} Our neural retargeter resolves these postural artifacts to provide a smooth and physically plausible global pose. \textit{Row 4 Coarse with Refinement:} This stage further optimizes end effector positioning to eliminate residual mesh penetrations and achieve precise contact alignment.}
    \label{fig:retargeting_qualitative}
\end{figure}

\noindent \textbf{Qualitative Analysis.} We visually compare the retargeted sequences against the ground truth in Figure \ref{fig:retargeting_qualitative}.
The Mapping+IK baseline struggles to maintain consistent spatial relationships and produces unnaturally distorted postures.
Conversely, our Coarse retargeter resolves these artifacts, producing a globally coherent pose that respects the physical constraints of the simulator.
The Coarse with Refinement demonstrates precise contact alignment by subtly adjusting the body to perfectly support the object geometry without penetration.

\section{Conclusion}
We presented a Policy as Data framework designed to bridges the representation gap between physics simulators and kinematic generative models and leverages expert policy trained in simulators as a data engine.
By utilizing a coarse-to-fine retargeting module to process the trajectory generated by reinforcement learning experts, our approach enables training generalizable diffusion models to synthesize physically accurate, multi-stage interactions.
Despite these advantages, scaling the framework to entirely novel task distributions remains constrained by the computational efficiency and reward engineering required to train task-specific expert policies. 
A promising direction for future work is to leverage the learned diffusion model as a prior to guide the RL training for new tasks.
Ultimately, this work establishes a robust foundation for generalizable and physically grounded motion synthesis.

%
%
\bibliographystyle{splncs04}
\bibliography{main}
\appendix
\begin{center}
    \Large \bfseries 
    Policy-as-Data: Learning Generalizable HOI Diffusion Models from Simulated Physics \\
    Supplementary
\end{center}

\section{Expert Policy Training}

This section details the comprehensive mathematical formulation and optimization strategy for the reinforcement learning experts introduced in Section \ref{sec:data_synthesis} of the main text.
The objective is to train a control policy $\pi$ via reinforcement learning (RL) that enables the character to perform functional tasks while maintaining realistic movement.
To ensure that synthesized interactions are both functional and anthropomorphically natural, we adopt a dual-component reward that decouples task specification from motion style~\cite{peng2021amp, InterPhysHassan2023}. Let $s_t$ denote the character-environment state at time step $t$, and let $a_t$ denote the action applied by the policy. The reward is defined as:
\begin{equation}
 r(s_t,a_t,s_{t+1},g)
 =
 w_G\, r_G(s_t,a_t,s_{t+1},g)
 +
 w_S\, r_S(s_t,s_{t+1}),
 \tag{\ref{eq:amp_reward}}
\end{equation}
where $r_G$ specifies t
where $r_G$ specifies the \emph{task} objective, e.g., reaching an object or transporting it to a target waypoint $g$, and $r_S$ specifies a \emph{style} objective that encourages human-like transitions. 

To provide a task-agnostic motion prior, $r_S$ employs a discriminator $D_\psi$ trained to distinguish simulated policy transitions $(s_t, s_{t+1})$ from real transitions in an unstructured MoCap dataset $\mathcal{M}$. $D_\psi$ is optimized via the standard adversarial objective:

\begin{equation}
\mathcal{L}_{\psi} = \mathbb{E}_{(s_t, s_{t+1}) \sim d^{\mathcal{M}}} \left[-\log D_{\psi}(s_t, s_{t+1})\right] + \mathbb{E}_{(s_t, s_{t+1}) \sim d^{\pi}} \left[-\log(1 - D_{\psi}(s_t, s_{t+1}))\right],
\end{equation}
where $d^{\mathcal{M}}$ represents the distribution of state transitions in the MoCap dataset and $d^{\pi}$ denotes the state-transition distribution induced by the policy.


The policy is then rewarded for fooling the discriminator, which encourages natural human kinematics across diverse tasks without rigidly tracking specific reference clips. 
The style reward for a given transition is then defined as:
\begin{equation}
r_S(s_t, s_{t+1}) = -\log(1 - D_{\psi}(s_t, s_{t+1})).
\label{eq:style_reward}
\end{equation}

The policy $\pi(a_t|s_t, g)$ is trained to maximize the expected cumulative discounted reward over a distribution of goals $g\sim p(g)$ and trajectories $\{s_0, a_0, s_1, a_1, ...\}$ induced by the policy:
\begin{equation}
J(\pi) = \mathbb{E}_{g \sim p(g), (s_t, a_t, s_{t+1}) \sim \pi} \left[ \sum_{t=0}^{T} \gamma^t r(s_t, a_t, s_{t+1}, g)\right],
\label{eq: RL objective}
\end{equation}
where $\gamma \in [0, 1)$ is the discount factor.
This formulation encourages the agent to complete tasks successfully while simultaneously adopting the kinematic styles found in the reference MoCap data.

\section{Contact Aware Refinement}
This section provides the detailed mathematical formulation for the contact aware refinement stage introduced in Section \ref{sec:retargeting} of the main text.

\noindent\textbf{Geometric Formulation and Signed Distance Field.}
We instantiate the object geometry at each frame as an oriented box and evaluate all geometric losses through its signed distance field SDF. Given box rotation $\mathbf{R}_t \in SO(3)$ translation $\mathbf{o}_t \in \mathbb{R}^3$ and half size vector $\mathbf{b}\in\mathbb{R}^3$ the box local coordinate of a 3D point $\mathbf{p}$ is
\begin{equation}
\tilde{\mathbf{p}}_t = \mathbf{R}_t^\top (\mathbf{p} - \mathbf{o}_t),
\end{equation}
and the SDF is written as
\begin{equation}
\phi(\mathbf{p}, t) = \mathrm{SDF}_{\mathrm{box}}(\tilde{\mathbf{p}}_t, \mathbf{b}),
\end{equation}
where $\phi(\mathbf{p}, t)>0$ denotes points outside the box and $\phi(\mathbf{p}, t)<0$ denotes penetration. In our final setting $\{\mathbf{R}_t, \mathbf{o}_t\}_{t=1}^T$ correspond to the frozen object trajectory $\hat{\boldsymbol{x}}^o_t$ and are kept fixed during refinement.

\noindent\textbf{Optimization Space and Temporal Masking.}
As stated in the main text the human state $\hat{\boldsymbol{x}}^{h}_t$ is refined to improve contact. The implementation achieves this palm centered refinement by optimizing a small upper body subset of the SMPL X pose parameters $\boldsymbol{\theta}_t$ extracted from $\hat{\boldsymbol{x}}^{h}_t$ namely the left and right collar shoulder elbow and wrist joints. Let $\hat{\boldsymbol{\theta}}_t$ be the retargeted initialization and $\Delta\boldsymbol{\theta}_t$ the optimized pose offset on these joints. Then the refined pose is
\begin{equation}
\boldsymbol{\theta}_t = \hat{\boldsymbol{\theta}}_t + m_t^{\mathrm{opt}} \odot \Delta\boldsymbol{\theta}_t,
\end{equation}
where $m_t^{\mathrm{opt}}\in\{0,1\}$ is a binary optimization mask. The body shape parameters hand pose parameters root translation and root orientation remain fixed.

Crucially we define the contact periods using the simulator provided contact indicators $c_t$. We derive specific active contact masks $m_{t}^{\mathrm{cont}}$ directly from $c_t$ and spatial proximity. In practice strict contact is identified using a distance threshold of $\tau_{\mathrm{cont}}=0.02$ m while a close range mask $m_{t}^{\mathrm{close}}$ is defined using $\tau_{\mathrm{close}}=0.12$ m. To avoid unstable corrections from severely invalid frames we only refine frames whose initial hand object SDF values are above a lower correction bound of $-0.15$ m. The final optimization mask $m_t^{\mathrm{opt}}$ is obtained by dilating these simulator guided masks by $4$ frames in time to ensure smooth transitions.

\noindent\textbf{Contact Enforcement Objectives.}
The contact term $E_{\mathrm{cont}}$ from the main text is practically expanded into surface attraction close range attraction no slip regularization and palm aware face consistency. For the interaction vertices $\mathcal{V}_{\mathrm{contact}}$ sampled from the palms we define proximity weights
\begin{equation}
w_{j,t} = \exp\left[-\frac{|\phi(\mathbf{v}_j(t), t)|}{\sigma}\right],
\end{equation}
where we set $\sigma=0.01$ m. The surface attraction loss on strict contact frames is
\begin{equation}
E_{\mathrm{surf}} = \frac{1}{|{\mathcal T}_{\mathrm{cont}}|} \sum_{t\in{\mathcal T}_{\mathrm{cont}}} \frac{\sum_{j\in\mathcal{V}_{\mathrm{contact}}} w_{j,t}\,\phi(\mathbf{v}_j(t), t)^2}{\sum_{j\in\mathcal{V}_{\mathrm{contact}}} w_{j,t}},
\end{equation}
where ${\mathcal T}_{\mathrm{cont}}$ is the set of frames where $c_t$ indicates active contact. We apply a similar weighted SDF loss $E_{\mathrm{close}}$ on close frames to avoid hovering palms. The no slip term enforces temporal consistency of contacted hand vertices in box local coordinates
\begin{equation}
E_{\mathrm{slip}} = \frac{\sum_{t=1}^{T-1}\sum_{j\in\mathcal{V}_{\mathrm{contact}}} \bar{w}_{j,t}\, \left\| \tilde{\mathbf{v}}_j(t+1)-\tilde{\mathbf{v}}_j(t) \right\|_2^2}{\sum_{t=1}^{T-1}\sum_{j\in\mathcal{V}_{\mathrm{contact}}}\bar{w}_{j,t}},
\end{equation}
where $\tilde{\mathbf{v}}_j(t)$ denotes the box local coordinate and $\bar{w}_{j,t}=\frac{1}{2}\left[w_{j,t}+w_{j,t+1}\right]$.

To aggregate these objectives the corresponding weights are set to $\lambda_{\mathrm{cont}}=3.0$ for surface attraction $\lambda_{\mathrm{close}}=1.0$ for close range attraction and $\lambda_{\mathrm{noslip}}=1.0$ for the no slip term. We further use palm face consistency weights of $\lambda_{\mathrm{plane}}=1.0$ $\lambda_{\mathrm{orient}}=0.5$ and a fallback palm attraction weight of $\lambda_{\mathrm{palm}}=0.5$ to encourage stable palm object attachment. The final contact objective is formulated as
\begin{equation}
E_{\mathrm{cont}} = \lambda_{\mathrm{cont}} E_{\mathrm{surf}} + \lambda_{\mathrm{close}} E_{\mathrm{close}} + \lambda_{\mathrm{noslip}} E_{\mathrm{slip}} + \lambda_{\mathrm{palm}} E_{\mathrm{palm}} + \lambda_{\mathrm{plane}} E_{\mathrm{plane}} + \lambda_{\mathrm{orient}} E_{\mathrm{orient}}.
\end{equation}

\noindent\textbf{Penetration Smoothness and Prior Penalties.}
To ensure the refined motion remains physically plausible and free of jitter we explicitly define the smoothness term $E_{\mathrm{sm}}$ introduced in the main text as an $\ell_2$ penalty on the joint accelerations of the optimized offsets
\begin{equation}
E_{\mathrm{sm}} = \sum_{t=2}^{T-1} \left\| \Delta\boldsymbol{\theta}_{t+1} - 2\Delta\boldsymbol{\theta}_t + \Delta\boldsymbol{\theta}_{t-1} \right\|_2^2.
\end{equation}
The penetration term $E_{\mathrm{pen}}$ is
\begin{equation}
E_{\mathrm{pen}} = E_{\mathrm{hand\_pen}} + E_{\mathrm{body\_pen}} + E_{\mathrm{arm\_clr}}.
\end{equation}
Here $E_{\mathrm{hand\_pen}}$ penalizes negative SDF values on $\mathcal{V}_{\mathrm{contact}}$ while $E_{\mathrm{body\_pen}}$ penalizes penetration of a subset of nearest to surface body vertices and $E_{\mathrm{arm\_clr}}$ penalizes elbow to wrist segments that breach a safety margin. In practice the hand penetration depth is clamped at $0.10$ m the hand penetration weight is set to $8.0$ the body penetration weight is set to $3.0$ and $2000$ body vertices nearest to the object surface are used for the coarse body penetration term.

Finally the reconstruction and prior terms are implemented as an $\ell_2$ penalty on the optimized pose offsets keeping the refined motion close to the retargeted initialization. We set the pose regularization weight to $0.1$ and the pose smoothness weight to $3.0$. Optimization is performed using Adam with a learning rate of $5\times 10^{-3}$. During the first stage of 200 iterations we optimize strictly for penetration resolution smoothness and pose regularization with all contact objectives disabled. In the second stage of 400 iterations we activate the full $E_{\mathrm{cont}}$ objective to securely anchor the hands to the object geometry while maintaining the collision free state achieved in the first stage.

\section{Dataset}To train our generative model, we utilize a hybrid dataset consisting of real-world motion capture and synthetic trajectories generated via our Policy-as-Data framework.\noindent \textbf{Real-World Data ($\mathcal{D}_{\text{real}}$):}We utilize OMOMO~\cite{li2023object}, a large-scale motion capture dataset containing diverse human-object interactions. Specifically, it features full-body manipulation sequences where subjects perform everyday tasks, such as pushing, pulling, carrying, and repositioning various household objects. These sequences capture complex human-object coordination, providing our framework with a robust prior for natural postural synthesis.\noindent \textbf{Simulator Data ($\mathcal{D}_{\text{sim}}$):}To provide the generative model with diverse, physically grounded functional skills, we generate raw rigid-body trajectories across three distinct interaction tasks. To support language-conditioned generation, the text prompts associated with these synthetic trajectories follow standardized templates based on the task category.\textbf{(1) Carry Tasks:} This category encompasses complete, multi-stage behaviors including approaching, picking up, carrying, and placing objects. We collect two subsets to encourage different types of generalization. First, to impart robust geometric adaptation, we generate 400 episodes (each lasting 300 frames) of carrying boxes with randomized sizes sampled uniformly from $[0.3, 0.4]$ meters. Moreover, the boxes are initialized at a random distance of 3 to 4 meters from the humanoid. The goal positions are sampled from a circle around the initial position of the object with a radius of 1.5 to 3 meters. Second, to ensure instance-level generalization across complex topologies, we replace the boxes with 30 distinct tables and 90 distinct chairs, collecting an additional 2048 episodes. The initialization distances and goal sampling strategies remain identical to those used for the boxes. All sequences in this category use the text prompt: "A person approaches a chair/box/table, picks it up, and places it in the designated location." \textbf{(2) Sit Tasks:} To further capture motions in diverse human-object interactions, we generate 1024 trajectories, each lasting 200 frames, of the humanoid interacting with and sitting on 30 different chair geometries. The associated text prompt for these episodes is: "A person approaches and sits on a chair."\textbf{(3) Climb Tasks:} To provide the model with dynamic locomotion and whole-body coordination skills, we collect 1024 trajectories, each lasting 200 frames, where the humanoid character climbs from the ground onto an elevated platform or a table. These sequences are annotated with the prompt: "A person climbs onto a platform/table."

\section{Implementation Details.}
We implement the full pipeline in PyTorch. The physics-based interaction trajectories are generated in Isaac Gym, which enables highly parallelized synthesis of human-object interaction data. The diffusion model is trained on eight NVIDIA RTX 4090 GPUs with mixed-precision training, while the retargeter follows the AvatarPoser training setup. More implementation details are provided in the Supplementary Material.

\noindent \textbf{Diffusion Model.}
Our diffusion model follows the implementation in CHOIS~\cite{li2024controllable}. It is a transformer-based conditional Gaussian diffusion model defined over motion windows of length 120. The predicted motion representation has dimension 216, including object translation (3), object rotation matrix (9), global human joint positions ($24\times 3$), and human joint rotations in 6D representation ($22\times 6$). When semantic contact channels are enabled, an additional 4 contact dimensions are appended. The object condition is represented using a $1024\times 3$ basis-point-set descriptor, which is flattened and encoded by a two-layer MLP with dimensions $3072 \rightarrow 512 \rightarrow 256$. The denoiser is a transformer decoder with model dimension 512, 4 decoder layers, 4 attention heads, and key/value dimensions of 256. A sinusoidal diffusion-timestep embedding is processed by an MLP and injected as an extra conditioning token. The model uses full self-attention over the entire temporal window. When language conditioning is enabled, text is encoded by CLIP ViT-B/32 and projected to the same 512-dimensional conditioning space. We use a cosine diffusion schedule with 1000 denoising steps and train the model with the \texttt{pred\_x0} objective and an $\ell_1$ reconstruction loss. Training uses Adam with learning rate $2\times 10^{-4}$, batch size 32, gradient accumulation over 2 steps, EMA decay 0.995, and a total of $8\times 10^5$ optimization steps.

\noindent \textbf{Coarse-to-Fine Retargeter.}
We designed our retargeter based on AvatarPoser~\cite{jiang2022avatarposer}. Unlike the original AvatarPoser which only takes the head and left-right hands as input, our retargeter takes a 270-dimensional sparse-motion descriptor extracted from the full 15-joint physics humanoid. To mitigate the scale and kinematic differences between the SMPL model and the physics humanoid, we first optimize a set of SMPL shape parameters $\beta$. This pre-optimization aligns the bone lengths (e.g., arms and legs) and the pelvis height of the SMPL model to match the specific kinematic attributes of the physics humanoid. For the network input, we concatenate local rotation in 6D, global rotation in 6D, linear velocity, and angular velocity for each joint, yielding $15\times(6+6+3+3)=270$ dimensions per frame. The network processes a temporal window of 40 frames using a linear input embedding of size $270 \rightarrow 256$, followed by a 3-layer Transformer encoder with 8 attention heads. Two lightweight MLP heads decode the final temporal features: one predicts the root orientation in 6D ($256 \rightarrow 256 \rightarrow 6$), and the other predicts the 21 body-joint rotations in 6D ($256 \rightarrow 256 \rightarrow 126$). The model supervises the last 3 frames of each window. During training, a differentiable SMPL-H body model is used for forward kinematics to recover 24 joints for positional supervision. Crucially, we apply the previously optimized $\beta$ parameters rather than the default zero shape vector during these kinematic computations to ensure accurate spatial alignment. We use an $\ell_1$ loss on root orientation, body-joint rotations, and joint positions, together with a velocity loss on predicted joint trajectories. The total loss is
\begin{equation}
\mathcal{L}{\mathrm{ret}} = 0.02\mathcal{L}{\mathrm{root}} + \mathcal{L}{\mathrm{pose}} + \mathcal{L}{\mathrm{jpos}} + 0.5\mathcal{L}_{\mathrm{vel}}.
\end{equation}
Training uses Adam with learning rate $10^{-4}$ and batch size 256. The learning rate is decayed by a factor of 0.5 at iterations 60k, 120k, 180k, 240k, 300k, and 360k.

\section{Evaluation Metrics}
To comprehensively assess generation quality, physical plausibility, and control adherence, we employ three sets of standard metrics following prior work~\cite{li2024controllable}:

\noindent \textbf{Condition Matching Metric:} This metric evaluates adherence to spatial constraints via the Euclidean distance between predicted and input object waypoints. We report the start position error ($T_s$), end position error ($T_e$), and intermediate position error ($T_{xy}$), all measured in centimeters.

\noindent \textbf{Interaction Quality Metric:} This metric assesses the physical fidelity of human-object contact. We report contact Precision ($C_{\text{prec}}$), Recall ($C_{\text{rec}}$), F1 score ($C_{F1}$), and overall contact percentage ($C_\%$). To measure physical violations, we report a penetration score ($P_{\text{hand}}$), calculated as the average penetration depth of interacting human vertices into the Signed Distance Field (SDF) of the object.

\noindent \textbf{Ground Truth Difference Metric:} This metric measures kinematic deviation from the reference motions. We compute the Mean Per-Joint Position Error (MPJPE), root translation error ($T_{\text{root}}$), and object position error ($T_{\text{obj}}$) via Euclidean distance (in cm). We also report the object orientation error ($O_{\text{obj}}$), which is measured using the Frobenius norm of the rotational difference.
\begin{table*}[t!]
    \centering
    \caption{\textbf{Data Proportion Ablation.} We evaluate generation performance under a strictly constant training budget where real motion capture samples are systematically replaced with our synthetic trajectories. The results demonstrate that increasing the synthetic data ratio steadily improves contact stability and interaction precision. This mathematically confirms that our simulator generated data provides a superior physical supervision signal while successfully preserving overall kinematic naturalness.}
    \label{tab:ablation_substitution}
    \resizebox{0.95\textwidth}{!}{
    \begin{tabular}{lccccccccc}
        \toprule
        & \multicolumn{5}{c}{Interaction Quality} & \multicolumn{4}{c}{Ground Truth Difference} \\
        \cmidrule(lr){2-6}\cmidrule(lr){7-10}
        Configuration & $C_{rec}\uparrow$ & $C_{prec}\uparrow$ & $C_{F1}\uparrow$ & $C_{\%}$ & $P_{hand}\downarrow$ & MPJPE$\downarrow$ & $T_{root}\downarrow$ & $T_{obj}\downarrow$ & $O_{obj}\downarrow$ \\
        \midrule
        100\% Real Baseline & 0.64 & \textbf{0.80} & 0.67 & 0.54 & \textbf{0.59} & \textbf{15.30} & \textbf{24.43} & \textbf{12.53} & \textbf{0.99} \\ 
        \ours 30\% Sim 70\% Real & 0.65 & 0.78 & 0.67 & 0.56 & 0.66 & 16.90 & 26.50 & 14.95 & 1.04 \\
        \ours 50\% Sim 50\% Real & 0.66 & 0.76 & 0.67 & \textbf{0.59} & 0.63 & 16.99 & 26.83 & 14.89 & 1.04 \\
        \ours 80\% Sim 20\% Real & \textbf{0.67} & 0.79 & \textbf{0.69} & 0.57 & 0.61 & 17.44 & 27.29 & 16.97 & 1.08 \\
        \bottomrule
    \end{tabular}
    }
\end{table*}

\section{Additional Ablation Studies}
\label{sec:supp_ablations}

To further investigate the intrinsic quality of our physics based synthetic data we conduct a data proportion ablation where we keep the total number of training samples constant but vary the ratio of real versus synthetic data. Unlike the experiments in the main text where $\mathcal{D}_{\text{sim}}$ is appended to the full $\mathcal{D}_{\text{real}}$ set here we maintain a constant training set size $N = |\mathcal{D}_{\text{real}}|$. We systematically replace portions of the real world MoCap data with our synthesized trajectories evaluating substitution ratios from 0\% Real baseline to 80\% synthetic composition.

As summarized in Table \ref{tab:ablation_substitution} the results reveal two critical insights. First even when the model is deprived of a significant portion of real world supervision the generation quality remains remarkably stable. Notably the contact stability $C_{\%}$ and interaction precision $C_{F1}$ improve as the synthetic ratio increases. This suggests that our simulator generated data provides a cleaner and more physically consistent supervision signal for interaction than raw MoCap which often contains capture noise or imprecise hand object alignments. Second while kinematic errors such as MPJPE show a marginal increase as real world samples are removed the model maintains high naturalness. This demonstrates that $\mathcal{D}_{\text{sim}}$ serves as a high fidelity proxy that effectively regularizes the diffusion model ensuring it learns functional interaction mechanics rather than merely memorizing specific kinematic patterns from the real world distribution.

\end{document}